\documentclass[11pt]{article}

\usepackage[margin=1in]{geometry}
\usepackage{amsmath,amssymb,amsthm,mathtools}
\usepackage[numbers,sort&compress]{natbib}
\usepackage{microtype}

\usepackage[dvipsnames]{xcolor}
\usepackage{hyperref}
\hypersetup{
  colorlinks=true,
  linkcolor=MidnightBlue,
  citecolor=OliveGreen,
  urlcolor=BrickRed
}

\newcommand{\softmax}{\operatorname{softmax}}
\newcommand{\KL}{\operatorname{KL}}
\newcommand{\R}{\mathbb{R}}
\newcommand{\one}{\mathbf{1}}
\newcommand{\angles}[1]{\langle #1 \rangle}
\newcommand{\IntDelta}{\operatorname{Int}\Delta}

\title{Softmax as a Lagrangian--Legendrian Seam}
\author{Christopher R. Lee--Jenkins}
\date{\today}

\begin{document}
\maketitle

\begin{abstract}
This note offers a first bridge from machine learning to modern differential geometry. We show that the logits-to-probabilities step implemented by softmax can be modeled as a geometric interface: two potential-generated, conservative descriptions (from negative entropy and log-sum-exp) meet along a Legendrian “seam” on a contact screen (the probability simplex) inside a simple folded symplectic collar. Bias-shift invariance appears as Reeb flow on the screen, and the Fenchel–Young equality/KL gap provides a computable distance to the seam. We work out the two- and three-class cases to make the picture concrete and outline “next steps” for ML: compact logit models (projective or spherical), global invariants, and connections to information geometry where on-screen dynamics manifest as replicator flows. 
\end{abstract}

\section{Introduction}
Bridging modern differential geometry with modern machine learning is delicate: the two fields organize knowledge differently. Differential geometry encodes structure through forms and flows (and proves invariance from those), while machine learning arranges modules around convex objectives, gradients, and probabilistic calibration. Yet there is a natural meeting point. Symplectic geometry has long provided the correct language for Hamiltonian (energy-conserving) systems \citep{Arnold1989,MarsdenRatiu1999}, and contact geometry, originating in Sophus Lie’s theory of contact transformations \citep{Lie1890} and developed in modern expositions such as \citet{Geiges2008}, models odd-dimensional, driven or thermodynamic evolution, including dissipative phenomena via contact Hamiltonian dynamics \citep{Bravetti2017}. 

Because contemporary ML pipelines routinely interleave conservative transforms (e.g., linear/residual maps) with normalization or projection steps (e.g., softmax, temperature scaling, probability calibration) that introduce divergence terms such as KL \citep{Goodfellow-et-al-2016,Amari2016}, it is natural to expect a symplectic–contact interface to appear. In this note we show that \emph{softmax} realizes precisely such an interface: two potential-generated (Lagrangian) descriptions meet on a contact screen along a Legendrian seam, with bias-shift invariance integrating the Reeb flow. The potentials come from the convex-dual pair (negative entropy, log-sum-exp), and the equalities reduce to the Fenchel--Young identity \citep{Rockafellar1970}. This note is addressed to readers fluent in machine learning techniques but not necessarily trained in differential geometry. Our aim is to describe, in the language of gradients and line integrals, a small piece of symplectic/contact geometry that explains how softmax functions as a seam between two conservative descriptions of the same data.

Two ideas suffice. First, a \emph{Lagrangian graph} here means nothing more exotic than the graph of a gradient of a scalar potential. If $F$ is a function, then $(x,\nabla F(x))$ is conservative: line integrals of $\nabla F$ depend only on endpoints. In our setting there are two potentials,
\[
\phi(y)=\sum_{i=1}^d y_i\log y_i \quad\text{(negative entropy)},\qquad
\phi^*(z)=\log\sum_{i=1}^d e^{z_i} \quad\text{(log-sum-exp)},
\]
and the familiar equalities
\[
y=\nabla\phi^*(z)=\softmax(z),\qquad
z=\nabla\phi(y)=\log y + c\,\one\ (c\in\R)
\]
say that logits and probabilities are related by gradients of dual potentials.

Second, a \emph{Legendrian} set lives on a hypersurface and is specified by a $1$-form. Concretely, a $1$-form $\alpha$ takes an infinitesimal change and returns a weighted line element (think “work”): if probability changes by $dy$, then
\[
\alpha(dy)=\sum_{i=1}^d z_i\,dy_i = z\cdot dy.
\]
The \emph{kernel} $\ker\alpha = \{ v \ | \ \alpha(v)=0\}$ consists of directions along which this work vanishes. A submanifold $S$ is Legendrian when every tangent vector to $S$ lies in $\ker\alpha$. There is also a canonical direction on the hypersurface, the \emph{Reeb} direction, characterized by $\alpha(R)=1$ and $d\alpha(R, \cdot)=0$; moving along $R$ changes coordinates without changing any measurement derived from $d\alpha$.

Softmax exposes exactly this structure. Probabilities live on the simplex $\Delta^{d-1}$, while logits live in $\R^d$ but only up to adding constants because $\softmax(z)=\softmax(z+c\one)$. On the product of these spaces there is a canonical $1$-form $\alpha=\sum_i z_i\,dy_i$ that records how logits weight infinitesimal probability changes. The seam where logits and probabilities agree (in the Fenchel–Young sense) is invisible to $\alpha$, hence Legendrian. The invariance $z\mapsto z+c\one$ is motion along the Reeb direction: it changes representatives of the same class of logits but leaves all measurements induced by $d\alpha$ unchanged.

To place this seam between two conservative descriptions, we thicken the simplex by one real coordinate $r$ transverse to it (we can take $r$ to be \emph{temperature}) and introduce the \emph{quadratic collar}
\[
\omega_q \;=\; dr\wedge \alpha \;+\; r^2\,d\alpha.
\]
For $r\neq 0$ this $2$-form is nondegenerate; at $r=0$ it loses rank in a controlled way—the \emph{fold}. On the two sides $r>0$ and $r<0$ the graphs of $d\phi^*$ and $d\phi$ appear as Lagrangian and meet along a Legendrian boundary on the fold. Verifying that one lies on the seam is a computation:
\[
\phi(y)+\phi^*(z)-\angles{z,y}\;=\;\KL\!\bigl(y\,\|\,\softmax(z)\bigr)\ \ge 0,
\]
with equality precisely when $y=\softmax(z)$. (Here, $\KL(y\|q)=\sum_i y_i\log(y_i/q_i)$ for $y,q$ on the simplex, nonnegative and zero exactly when $y=q$.)

\section{Screen, collar, and seam}\label{sec:one-section}
We now formalize the ideas of the previous section more thoroughly, providing a bridge between the geometry of logits and (folded) symplectic geometry. Let $\IntDelta$ be the interior of the simplex and let $\overline{\mathbb{R}^d}=\mathbb{R}^d/\langle \mathbf{1}\rangle$ denote logits modulo constant shifts. Define the \emph{screen}
\[
\mathcal{Z}=\{(y,[z]) : y\in\IntDelta,\ [z]\in \overline{\R^d}\}.
\]
Let the \emph{interface} $Z \subset \mathbb{R}^d$ project to $\mathcal{Z}$. On $Z$ consider the $1$-form
\begin{equation}\label{eq:alpha}
\alpha=\sum_{i=1}^d z_i\,dy_i,
\end{equation}
which is well-defined because $\sum_i dy_i=0$ along $\IntDelta$. 

Thicken the interface by one coordinate $r\in\R$ and set $\mathcal{M}=\R_r\times Z$ with quadratic collar
\begin{equation}\label{eq:omegaq}
\omega_q = dr\wedge \alpha + r^2\,d\alpha, 
\end{equation}
the hypersurface $Z=\{r=0\}$ is the \emph{fold}. 
The pair $(\alpha,d\alpha)$ endows $Z$ with a contact structure in the standard sense; the flow $z\mapsto z+c\one$ integrates the Reeb vector field, making bias-shift invariance intrinsic to the screen.

Two potential-generated graphs occupy the sides of the collar:
\[
L_{+}=\{(r,y,z) : r>0,\ y=\nabla\phi^*(z)\},\qquad
L_{-}=\{(r,y,z) : r<0,\ z\in\nabla\phi(y)\}.
\]
Here $\nabla\phi^*(z)=\softmax(z)$ and $\nabla\phi(y)=\log y + c\,\one$ with class $[z]=[\log y]$. Each $L_{\pm}$ is Lagrangian in $(\mathcal{M}\setminus Z,\omega_q)$: on any slice $\{r=\pm\varepsilon\}$ the restriction is the graph of an exact differential in the exact symplectic manifold $\bigl(\{r=\pm\varepsilon\},\,d\theta_{\pm\varepsilon}\bigr)$ with primitive $\theta_{\pm\varepsilon}=(\pm\varepsilon)^2\alpha$.

The \emph{seam} on the fold is the set
\[
\Gamma=\{(0,y,[z])\in Z : y=\nabla\phi^*(z)\}.
\]
It is Legendrian: $\alpha$ annihilates all tangents to $\Gamma$, and both $L_{+}$ and $L_{-}$ close onto $\Gamma$ as $r\to 0^\pm$. The defining condition for $\Gamma$ is the equality case of Fenchel–Young,
\[
\angles{z,y}=\phi(y)+\phi^*(z)\iff y=\nabla\phi^*(z),
\]
and the corresponding nonnegativity
\[
\phi(y)+\phi^*(z)-\angles{z,y}=\KL\!\bigl(y\,\|\,\softmax(z)\bigr)\ge 0
\]
provides a quantitative, readily computed distance from any $(z,y)$ to the seam.

Finally, since $\softmax(z)=\softmax(z+c\one)$, inverting softmax requires a gauge (a choice of a constant difference direction between logits). A canonical choice is the zero-mean gauge
\[
\Pi(y)=\Bigl[\log y - \frac{1}{d}\Big(\sum_{i=1}^d \log y_i\Big)\one\Bigr]\in \R^d.
\]
The map $y\mapsto (r,y,\Pi(y))$ with $r<0$ parametrizes $L_{-}$ near the fold and exhibits how the construction passes back from probabilities to an equivalence class of logits; different gauges translate along the Reeb direction on $Z$.

\section{Examples}\label{sec:examples}

This section makes the construction concrete in the two- and three-class cases. Throughout, \(\phi(y)=\sum_i y_i\log y_i\) and \(\phi^*(z)=\log\sum_i e^{z_i}\), so that \(y=\nabla\phi^*(z)=\softmax(z)\) and \(z=\nabla\phi(y)=\log y + c\,\one\).

\subsection*{Two classes (\(d=2\)): sigmoid on an interval}
Write logits \(z=(a,b)\) and probabilities \(y=(p,1-p)\). The softmax reduces to
\[
p=\frac{e^{a}}{e^{a}+e^{b}}=\sigma(\Delta),\qquad \Delta:=a-b,
\]
with \(\sigma(t)=\frac{1}{1+e^{-t}}\). Bias-shift invariance \(z\mapsto z+c\one\) leaves \(\Delta\) unchanged. A canonical gauge is the zero-mean representative
\[
[z]=\Bigl(\tfrac{\Delta}{2},-\tfrac{\Delta}{2}\Bigr),\qquad
\Pi(y)=\Bigl(\tfrac12\log\frac{p}{1-p},-\tfrac12\log\frac{p}{1-p}\Bigr).
\]
The screen \(\mathcal{Z}\) is the open interval \(p\in(0,1)\) paired with the class \([z]\); its \(1\)-form is
\[
\alpha \;=\; z_1\,dp + z_2\,d(1-p) \;=\; (z_1 - z_2)\,dp \;=\; \Delta\, dp.
\]

\begin{figure}[t]
  \centering
  \includegraphics[scale=.85]{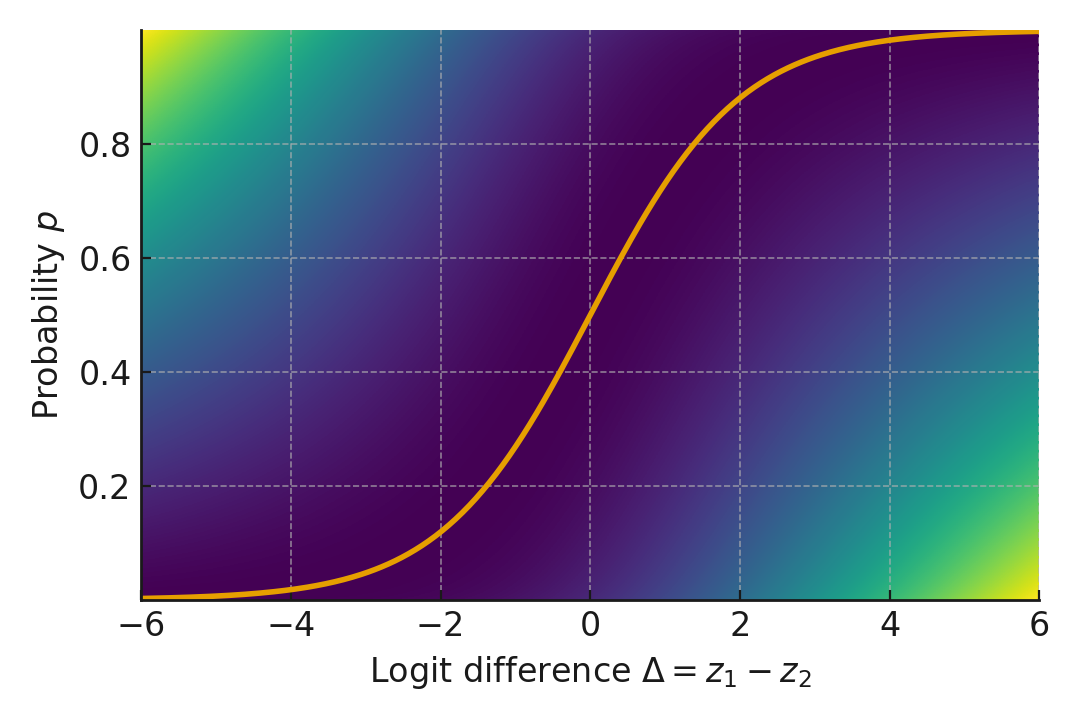}
  \caption{Two-class seam $p=\sigma(\Delta)$ (thick curve) on the $(\Delta,p)$-plane. 
  The background shows the Fenchel--Young gap 
  $\phi(y)+\phi^*(z)-\langle z,y\rangle=\mathrm{KL}\!\bigl(y\,\|\,\mathrm{softmax}(z)\bigr)$, 
  which vanishes exactly on the seam.}
  \label{fig:two-class}
\end{figure}

\begin{figure}[t]
  \centering
  \includegraphics[scale=.6]{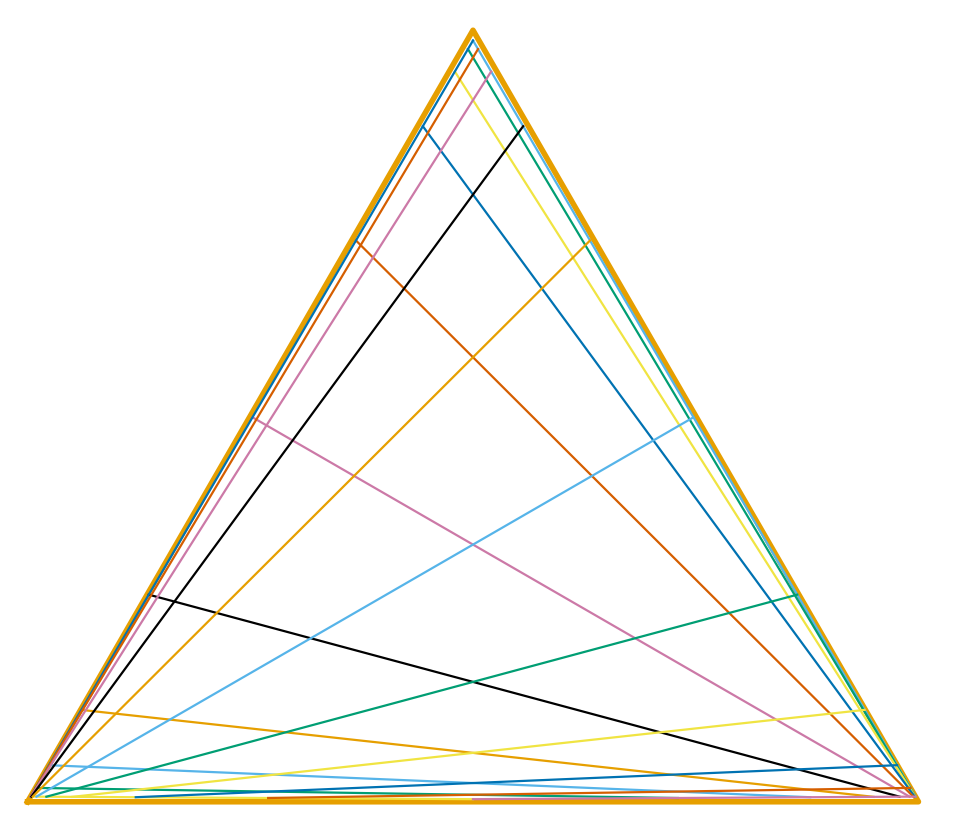}
  \caption{Three-class softmax: image in the probability simplex $\Delta^2$ of a rectangular grid 
  in centered-logits coordinates $(z_1\!-\!z_3,\;z_2\!-\!z_3)$. 
  Only logit differences are visible on the screen, reflecting bias-shift (Reeb) invariance.}
  \label{fig:three-class}
\end{figure}

The seam \(\Gamma\subset \{r=0\}\) consists of pairs \(\bigl(p,[z]\bigr)\) with \(p=\sigma(\Delta)\) (equivalently \(\Delta=\log\frac{p}{1-p}\)). The Fenchel--Young gap is the familiar binary KL divergence
\[
\phi(y)+\phi^*(z)-\angles{z,y}
= p\log\frac{p}{\sigma(\Delta)} + (1-p)\log\frac{1-p}{1-\sigma(\Delta)} \;\ge 0,
\]
which vanishes exactly on \(\Gamma\). This provides an immediate numerical criterion for being on the seam. The collar \(2\)-form is
\[
\omega_q \;=\; dr\wedge \alpha + r^2\,d\alpha \;=\; dr\wedge(\Delta\,dp) + r^2\, d\Delta\wedge dp,
\]
nondegenerate for \(r\neq 0\). The Lagrangian graphs are simply
\[
L_{+}=\{(r>0,\,p=\sigma(\Delta),\,[z])\},\qquad
L_{-}=\{(r<0,\, [z]=[\log p - \log(1-p),\,-(\log p - \log(1-p))])\},
\]
meeting along \(\Gamma\) at \(r=0\). Motion \(z\mapsto z+c\one\) translates along the Reeb direction on the screen and does not affect \(p\).

\subsection*{Three classes (\(d=3\)): triangle simplex and centered logits}
Write \(y=(u,v,1-u-v)\) with \(u,v>0\) and \(u+v<1\). Choose the centered gauge
\[
\tilde z \;=\; z - \tfrac13(z_1+z_2+z_3)\,\one \in \overline{\R^3},
\]
so \(\tilde z_1+\tilde z_2+\tilde z_3=0\). Softmax reads
\[
y_i \;=\; \frac{e^{\tilde z_i}}{e^{\tilde z_1}+e^{\tilde z_2}+e^{\tilde z_3}},\qquad i=1,2,3.
\]
The inverse (pass-back through the fold) in this gauge is
\[
\Pi(y) \;=\; \log y - \tfrac13\Bigl(\sum_{i=1}^3 \log y_i\Bigr)\one,
\]
which is well defined for \(y\in \IntDelta\). On the screen the \(1\)-form \(\alpha=\sum_i z_i\,dy_i\) becomes, in \((u,v)\)-coordinates,
\[
\alpha \;=\; z_1\,du + z_2\,dv + z_3\,d(1-u-v)
\;=\; (z_1-z_3)\,du + (z_2-z_3)\,dv,
\]
which depends only on logit \emph{differences} and is thus invariant under \(z\mapsto z+c\one\) (the Reeb flow). The seam \(\Gamma\subset\{r=0\}\) consists of pairs \(\bigl((u,v,1-u-v),[\tilde z]\bigr)\) with \(y=\softmax(\tilde z)\), or equivalently \([\tilde z]=[\log y]\) in the centered gauge. The KL expression
\[
\phi(y)+\phi^*(z)-\angles{z,y}
= \sum_{i=1}^3 y_i \log\frac{y_i}{\softmax(z)_i}\ \ge 0
\]
vanishes identically on \(\Gamma\) and is strictly positive off the seam. The quadratic collar
\[
\omega_q \;=\; dr\wedge\alpha + r^2\,d\alpha 
\;=\; dr\wedge\bigl((z_1-z_3)\,du + (z_2-z_3)\,dv\bigr)
\;+\; r^2\, d(z_1-z_3)\wedge du \;+\; r^2\, d(z_2-z_3)\wedge dv
\]
is nondegenerate for \(r\neq 0\). The two potential-generated graphs
\[
\begin{aligned}
L_{+}&=\{(r>0,\, y=\nabla\phi^*(z)=\softmax(z))\},\\
L_{-}&=\{(r<0,\, z\in\nabla\phi(y)=\log y + c\,\one)\}
\end{aligned}
\]
occupy the sides of the collar and close onto the seam \(\Gamma\) at \(r=0\).

\section{Conclusion}
Information-geometric approaches view the simplex as a Riemannian manifold (Fisher/Shahshahani), so dynamics constrained to the simplex appear as replicator flows—the continuous-time limit of multiplicative-weights updates \citep{Amari2016}. In this language, the \emph{screen} dynamics highlighted here are exactly replicator dynamics, and recent work formalizes their role in next-token prediction: the output distribution evolves inside the simplex by a replicator flow and converges to a softmax equilibrium \citep{LeeJenkins2025}. 

Our contribution is complementary: near the logits-to-probabilities step, two potential-generated graphs (of $d\phi$ and $d\phi^*$) occupy a quadratic symplectic collar and glue along a Legendrian seam on a contact screen, with bias-shift invariance integrating the Reeb flow. What we present is also \emph{local}. To leverage the full power of symplectic methods (global invariants, compactness arguments, and long-time dynamics), one would like a \emph{compact} ambient model for logit space—e.g., working modulo scale and mean so that logits descend to a projective space, or after normalization to a sphere. Such compactifications promise global versions of the local statements proved here, and open the door to studying screen–ambient interactions with the same structural control that symplectic geometry affords in classical mechanics.

\bibliographystyle{plainnat}

\end{document}